\pdfoutput=1

\documentclass[11pt]{article}

\usepackage[final]{acl}

\usepackage{times}
\usepackage{soul}
\usepackage{threeparttable}
\usepackage{subcaption}
\usepackage{latexsym}
\usepackage{graphicx}
\usepackage{hyperref}
\usepackage{amsmath}
\usepackage{times}
\usepackage{latexsym}
\usepackage{natbib}
\usepackage{enumerate}
\usepackage{amssymb}
\usepackage{multirow}
\usepackage{multicol}
\usepackage{enumitem}
\usepackage{booktabs}
\usepackage{enumitem}
\usepackage{balance}
\usepackage{natbib}
\usepackage{amssymb}
\usepackage{pifont}
\usepackage{blindtext}
\usepackage{tabularx}
\usepackage{xcolor}
\usepackage{colortbl}
\usepackage{makecell}

\newcolumntype{s}{>{\hsize=.12\hsize}X}
\newcolumntype{m}{>{\hsize=.3\hsize}X}
\usepackage[all]{nowidow}
\usepackage{microtype}
\usepackage{alltt}
\usepackage{tcolorbox}
\tcbuselibrary{skins}
\usepackage{enumitem}
\usepackage{tikz}
\usepackage{siunitx}
\usetikzlibrary{trees, positioning}

\newcolumntype{T}{S[table-format=1.3, table-space-text-post=\textsuperscript{***}]}

\definecolor{customorange}{HTML}{ffa602}
\definecolor{customblue}{HTML}{6394ee}
\definecolor{lightorange}{HTML}{ffd580} 
\definecolor{lightblue}{HTML}{a7c7f2}   
\definecolor{darkred}{RGB}{139,0,0}
\definecolor{darkblue}{RGB}{0,0,139}

\setstcolor{red}
\definecolor{ForestGreen}{rgb}{0.13, 0.55, 0.13}

\definecolor{VibrantBlue}{rgb}{0.0, 0.2, 1.0}

\definecolor{FireBrick}{rgb}{0.7, 0.13, 0.13}

\usepackage[T1]{fontenc}

\usepackage[utf8]{inputenc}

\usepackage{microtype}

\usepackage{inconsolata}

\usepackage{graphicx}

\title{\textsc{SubData}: Bridging Heterogeneous Datasets to Enable Theory-Driven Evaluation of Political and Demographic Perspectives in LLMs}

\author{Pietro Bernardelle$^1$ \ \ \ Leon Fröhling$^2$ \ \ \ Stefano Civelli$^1$ \ \ \ Gianluca Demartini$^1$ \\ 
$^1$~The University of Queensland, Australia \ \ \ \\$^2$~GESIS - Leibniz Institute for the Social Sciences, Germany \\
\small{\texttt{\{p.bernardelle,s.civelli,g.demartini\}@uq.edu.au}} \ \ \ \small{\texttt{leon.froehling@gesis.org}}}

\begin{document}
\maketitle
\begin{abstract}
As increasingly capable large language models (LLMs) emerge, researchers have begun exploring their potential for subjective tasks. While recent work demonstrates that LLMs can be aligned with diverse human perspectives, evaluating this alignment on downstream tasks (e.g., hate speech detection) remains challenging due to the use of inconsistent datasets across studies. To address this issue, in this resource paper we propose a two-step framework: we \textbf{(1)} introduce \textsc{SubData}, an open-source Python library\footnote{\url{https://github.com/Subdata-Library/Subdata/}}\textsuperscript{,}\footnote{\url{https://pypi.org/project/subdata/}} designed for standardizing heterogeneous datasets to evaluate LLMs perspective alignment; and \textbf{(2)} present a theory-driven approach leveraging this library to test how differently-aligned LLMs (e.g., aligned with different political viewpoints) classify content targeting specific demographics. \textsc{SubData}'s flexible mapping and taxonomy enable customization for diverse research needs, distinguishing it from existing resources. We illustrate its usage with an example application and invite contributions to extend our initial release into a multi-construct benchmark suite for evaluating LLMs perspective alignment on natural language processing tasks.  
\end{abstract}

\section{Introduction}
\label{s:intro}
\begin{figure}[t]
    \centering
    \includegraphics[width=\linewidth]{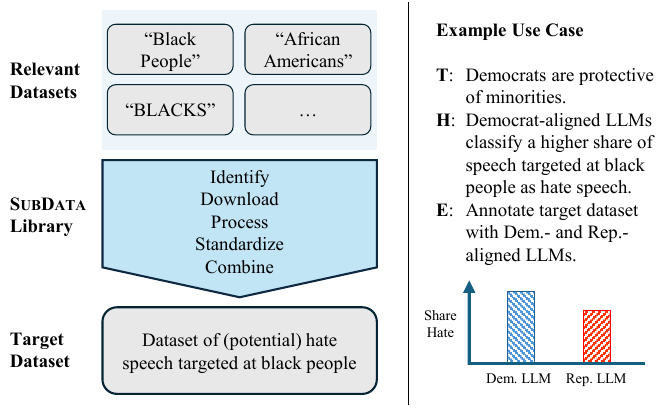}
    \caption{Overview of our proposed evaluation framework. \textsc{SubData} consolidates instances from diverse datasets into a unified resource. To assess LLM alignment with human perspectives from the combined dataset, we propose a workflow that tests theory-derived (\textbf{T}) hypotheses (\textbf{H}) through controlled experiments (\textbf{E}), measuring how accurately LLMs reflect viewpoints of different demographic and ideological groups.}
    \label{fig:f1}
\end{figure}

The ever-increasing capabilities of large language models (LLMs) have enabled them to capture increasingly nuanced human perspectives \cite{brown2020language, bommasani2021opportunities}. Researchers have begun exploring their potential for subjective tasks, with particular focus on ``perspective alignment''—the ability of models to reflect diverse human viewpoints across different contexts \cite{durmus2023towards, kirk2024prism}. Ensuring robust evaluation of this alignment is crucial as LLMs increasingly mediate information access and influence decisions in socially sensitive domains where human perspectives naturally differ \cite{blodgett2020language, weidinger2021ethical, khamassi2024strong}.

Recent work has explored how well LLMs can represent diverse human perspectives using two different approaches. The first approach examines whether models accurately predict how individuals \cite{argyle2023outofone} or groups \cite{santurkar2023whose} would respond to surveys, a task \citet{sorensen2024position} describe as \textit{distributional pluralism}. The second investigates whether aligned LLMs consistently reflect broader viewpoints across tasks \cite{feng2023pretraining, agiza2024politune, chen2024susceptible, haller2024opiniongpt, he2024community}, aligning with what \citet{sorensen2024position} term \textit{steerable pluralism}.

Survey prediction provides a natural evaluation setting: models’ outputs—generated either through fine-tuning or persona-conditioning to represent specific perspectives—can be compared against authentic survey responses from individuals or subpopulations. Because such datasets contain demographic information and corresponding answers, they create a clear benchmark: a well-aligned LLM should produce distributions that closely resemble real responses and can be evaluated using standard divergence or accuracy measures \cite{sorensen2024position}.

The broader challenge of task-independent alignment has inspired various evaluation methodologies. Political alignment studies by \citet{agiza2024politune} and \citet{chen2024susceptible} use the Political Compass Test (PCT)—a widely used questionnaire for mapping political beliefs along economic and social axes—to verify whether models aligned to specific ideologies position themselves appropriately on the PCT map. \citet{he2024community} compare model answers to multiple-choice questions against positions expressed by relevant subgroups. \citet{sorensen2024position} propose direct human annotations or reward models to measure whether generated responses correctly reflect specific attributes. More closely related to our conceptualization of alignment evaluation, \citet{haller2024opiniongpt} assess sentiment in open-ended generations when prompted about different demographics, while \citet{feng2023pretraining} examine how political alignment affects hate speech detection toward different targets.

Despite these efforts, evaluating how perspective-aligned LLMs perform on subjective classification tasks remains challenging \cite{zheng2024helpful}, largely due to the lack of standardized resources that enable consistent comparison across viewpoints \cite{alipour2024robustness}.
We address this gap by introducing a two-step framework that enables systematic evaluation of perspective-aligned language models.

\paragraph{(1) Dataset Standardization: \textsc{SubData}.}
We introduce \textsc{SubData}, an open-source Python library that collects and harmonizes heterogeneous datasets for subjective tasks. Unlike general repositories, it unifies inconsistent annotation schemes and demographic categorizations, allowing researchers to build consistent collections for their needs. Our initial release focuses on hate speech detection, integrating ten datasets with a unified taxonomy of target groups ($\S$\ref{s:subdata}, $\S$\ref{s:taxonomy}). In doing so, we do not host or redistribute the datasets themselves, but we reviewed their licenses to ensure that our use aligns with creators’ intentions of fostering hate speech research. Consistent with \citet{vidgen2020directions}, we further emphasize the need to handle such material responsibly, with attention to privacy, personal data, and potential online harms.

\paragraph{(2) Theory-Driven Hypothesis Testing.}
Building on these standardized datasets, we propose a theory-driven approach to evaluate alignment ($\S$\ref{s:theory}). As illustrated in Figure~\ref{fig:f1}, our framework follows a systematic process: researchers first formulate hypotheses (\textbf{H}) based on established social or political theory (\textbf{T}), then design experiments (\textbf{E}) to test whether differently-aligned models behave as expected. For instance, the workflow on the right illustrates testing whether Democrat-aligned LLMs classify more anti-Black content as hate speech than Republican-aligned ones, reflecting the popular hypothesis that Democrats prioritize minority protection theoretically derived by \cite{solomon2024illusory}. This framework enables quantitative measurement of alignment differences through controlled experimentation, and we further demonstrate its application in $\S$\ref{sec:exp-example}.

\paragraph{}
Our approach does not rely on subjective ground-truth labels; instead, it measures classification differences across models with distinct alignments, providing a direct lens on how perspective conditioning shapes downstream task behavior. While prior work has examined subjectivity in LLM annotation \cite{orlikowski2023ecological, beck2024sensitivity, giorgi2024human}, our framework extends this by systematically evaluating alignment effects in downstream applications.

\section{Related Work}
\subsection{LLMs Perspective Alignment}

Research on aligning LLMs with diverse human perspectives has followed two main approaches: fine-tuning models on perspective-specific data and using persona-based prompting.

Several studies have explored fine-tuning approaches for task-agnostic LLMs alignment. \citet{feng2023pretraining}, \citet{agiza2024politune} and \citet{chen2024susceptible} investigated how political alignment and data selection affect model biases and downstream tasks like hate speech detection. Similarly, \citet{haller2024opiniongpt} developed OpinionGPT by fine-tuning models on ideologically diverse data to represent explicit biases.

As an alternative to these resource-intensive post-training methods, persona-based prompting has emerged as a more efficient technique for task-specific perspective alignment. \citet{argyle2023outofone} showed that LLMs can accurately simulate survey responses across demographic groups, while \citet{ge2024scaling} and \citet{frohling2024personas} demonstrated how synthetic personas can diversify model outputs and annotations. Building on this, \citet{bernardelle2025political,bernardelle2025mapping} mapped persona-prompted LLMs onto the PCT compass, providing a large-scale analysis of how these personas impact the distribution of language models across political ideological space. Similarly, \citet{civelli2025impact} revealed how politically-aligned persona-conditioned LLMs influence hateful content detection.

\citet{orlikowski2025beyond} combined these approaches by fine-tuning models with socio-demographic attributes to represent individual annotators, finding that persona-based prompting barely improves the models' ability to predict individuals' annotations and that improvements from fine-tuning mainly come from demographic profiles serving as identifiers for individual annotators. \citet{liu2024evaluating} identified further limitations in this technique, showing that models struggle with ``incongruous personas'' and default to stereotypical stances when predicting responses for personas with contradicting traits. The conflicting evidence seen in the literature regarding the models' ability to consistently represent different subjective perspectives serves as further motivation to develop comprehensive resources for the evaluation of this type of LLMs perspective alignment. 

\subsection{Evaluating LLMs Perspective Alignment}

Evaluating alignment presents significant challenges, particularly for subjective tasks.

For survey response prediction, \citet{santurkar2023whose} and \citet{he2024community} compared model predictions against actual responses from specific demographic groups. \citet{castricato2025persona} built on the PRISM dataset \citep{kirk2024prism} to create a test bed for evaluating pluralistic alignment using preference pairs from personas sampled from census data.

For downstream tasks, \citet{zheng2024helpful} and \citet{giorgi2024human} assessed how personas affect model performance and biases in content classification. Despite these advances, evaluating perspective-aligned LLMs on subjective classification tasks remains challenging due to the lack of standardized resources that enable consistent comparison—a gap our proposed framework addresses.

\section{\textsc{SubData} Construction}
\label{s:subdata}

\begin{table*}[ht]
    \footnotesize
    \resizebox{\textwidth}{!}{
    \begin{tabular}{l|rrrrrrrrr|r}
        \toprule
        \textbf{Dataset} \textbackslash \hspace{0mm} \textbf{Category} & \textbf{age} & \textbf{disabled} & \textbf{gender} & \textbf{migration} & \textbf{origin} & \textbf{political} & \textbf{race} & \textbf{religion} & \textbf{sexuality} & \textbf{Dataset size}\\
        \midrule
        \citet{fanton2021human} & 0 (0) & 175 (1) & 560 (1) & 637 (1) & 0 (0) & 0 (0) & 301 (1) & 1,402 (2) & 465 (1) & 3,540 \\
        \citet{hartvigsen2022toxigen} & 0 (0) & 19,631 (1) & 19,563 (1) & 0 (0) & 62,458 (3) & 0 (0) & 80,979 (4) & 41,014 (2) & 21,344 (1) & 244,989 \\
        \citet{jigsaw2019} & 0 (0) & 18,602 (3) & 178,266 (4) & 0 (0) & 0 (0) & 0 (0) & 94,334 (5) & 132,734 (7) & 29,115 (4) & 453,051 \\
        \citet{jikeli2023antisemitism} & 0 (0) & 0 (0) & 0 (0) & 0 (0) & 0 (0) & 0 (0) & 0 (0) & 6,439 (1) & 0 (0) & 6,439 \\
        \citet{jikeli2023general} & 0 (0) & 0 (0) & 0 (0) & 0 (0) & 0 (0) & 0 (0) & 3,012 (3) & 2,315 (2) & 0 (0) & 5,327 \\
        \citet{mathew2021hatexplain} & 0 (0) & 153 (1) & 5,584 (2) & 1,701 (1) & 1,855 (2) & 0 (0) & 7,684 (5) & 6,106 (6) & 2,750 (4) & 25,833 \\
        \citet{rottger2021hatecheck} & 0 (0) & 510 (1) & 1,020 (2) & 485 (1) & 0 (0) & 0 (0) & 504 (1) & 510 (1) & 577 (1) & 3,606 \\
        \citet{sachdeva2022measuring} & 2,355 (4) & 1,801 (3) & 22,535 (5) & 5,473 (2) & 11,637 (2) & 0 (0) & 21,024 (7) & 12,461 (8) & 14,934 (4) & 92,220 \\
        \citet{vidgen2021introducing} & 41 (2) & 414 (3) & 689 (3) & 45 (2) & 164 (5) & 688 (7) & 397 (4) & 273 (4) & 472 (3) & 3,183 \\
        \citet{vidgen2021learning} & 23 (1) & 521 (1) & 3,630 (4) & 1,507 (2) & 862 (6) & 0 (0) & 3,881 (5) & 2,384 (2) & 1,437 (3) & 14,245 \\
        \midrule
        All Datasets & 2,419 (4) & 41,807 (3) & 231,847 (5) & 9,848 (4) & 76,976 (11) & 688 (8) & 212,116 (8) & 205,638 (8) & 71,094 (6) & 852,433 \\
        \bottomrule
    \end{tabular}
    }
    \caption{Overview of hate speech datasets in \textsc{SubData}, showing the number of instances and unique target groups (in parentheses) per target category. \textit{Note}: The ``All Dataset'' row reports the total unique target groups per category across all datasets. When the total equals the maximum from a single dataset (e.g., disabled: 3, matching \citet{jigsaw2019}’s 3), that dataset fully accounts for the category’s unique target groups. When the total exceeds the maximum (e.g., origin: 11, exceeding \citet{hartvigsen2022toxigen}’s 3), multiple datasets contribute distinct target groups.}
    \label{tab:datasets}
\end{table*}

\subsection{Dataset Selection Criteria}
Our approach to evaluating perspective alignment in LLMs necessitates datasets with specific characteristics suited for this analysis. We require datasets that address subjective constructs such as hate speech, toxicity, or abusive language—domains where human interpretations naturally diverge across demographic and ideological lines \cite{sap2021annotators}. This subjectivity is essential as it creates the interpretive space where different perspectives become measurable. Additionally, these datasets must provide explicit annotations identifying which specific demographic groups are targeted by the content (for example, specifying when content targets Jews, women, or immigrants), rather than merely indicating that some unspecified group was targeted. This granular targeting information is crucial because it enables us to test theory-driven hypotheses about how LLMs aligned with different perspectives might classify content targeting specific demographics differently.

\subsection{Data Collection Methodology}
Because of the lack of a single repository that stores and documents the properties of datasets, identifying the set of relevant datasets is an inherently difficult challenge. We therefore employed a multi-phase approach to identify suitable datasets. 

First, we leveraged our existing knowledge of hate speech detection literature to identify candidate datasets, drawing on our team's established expertise in this domain. Second, we examined existing repositories including \href{https://hatespeechdata.com/}{hatespeechdata.com} \cite{vidgen2020directions} and toxic-comment-collection \cite{risch2021data}, which provided structured access to multiple potentially relevant datasets. Third, we conducted systematic searches with keyword combinations of ``target[ed]'' and ``hate speech'' on scholarly databases to identify related literature that might present or reference additional resources. Finally, we individually assessed each dataset through manual verification to confirm it contained explicit target group annotations that satisfied our criteria. 

This process yielded ten datasets that meet our requirements. While we have striven to make our initial dataset collection comprehensive, we acknowledge that this collection is not exhaustive and that some relevant sources may have been overlooked. Rather than seeing this as a limitation, we consider it an opportunity to build a collaborative research community focused on annotation subjectivity. We actively encourage researchers to contact us with suggestions for additional datasets that satisfy our outlined criteria to be included in the library.

\subsection{Dataset Characteristics}
Table~\ref{tab:datasets} provides an overview of the datasets included in \textsc{SubData} so far, categorizing targets across nine demographic dimensions (age, disability, gender, migration, origin, political, race, religion, and sexuality). All target categories are organized according to the unified taxonomy we detail in $\S$\ref{s:taxonomy}, which standardizes the heterogeneous labels from original sources. This standardized categorization enables researchers to quickly identify suitable datasets for specific research questions regarding perspective alignment, highlighting both the strengths and limitations of current hate speech detection resources.

We would like to point out that the number of entries in some datasets of Table \ref{tab:datasets} may differ from those reported in the original publications because of our focus on targeted hate speech. When entries in source datasets had multiple targets in a single annotation (e.g., ``[bla, jew]''), we created separate instances for each target,  thereby increasing the number of entries. Conversely, we excluded entries without specific target groups (e.g., labeled as ``other''), resulting in datasets that sometimes contain fewer instances than the originals. We also deduplicate instances, removing repeated entry-target pairs even when these duplications might be intentional in the original dataset—such as in \citet{fanton2021human} where identical hate speech instances appear multiple times with different counterspeech responses. Since our research focuses specifically on targeted hate speech, we treat these as functional duplicates.

\section{\textsc{SubData} Unified Taxonomy}
\label{s:taxonomy}
Following our dataset selection and collection methodology, \textsc{SubData} implements a standardized taxonomy that addresses the inconsistencies in how target groups are labeled across hate speech datasets. This allows to leverage the systematic evaluation framework described in $\S$\ref{s:theory} by creating consistency across disparate data sources. Additional details on \textsc{SubData}'s functionalities are provided in Appendix~\ref{s:library}.

\begin{figure*}[t]
    \centering
    \includegraphics[width=1\linewidth]{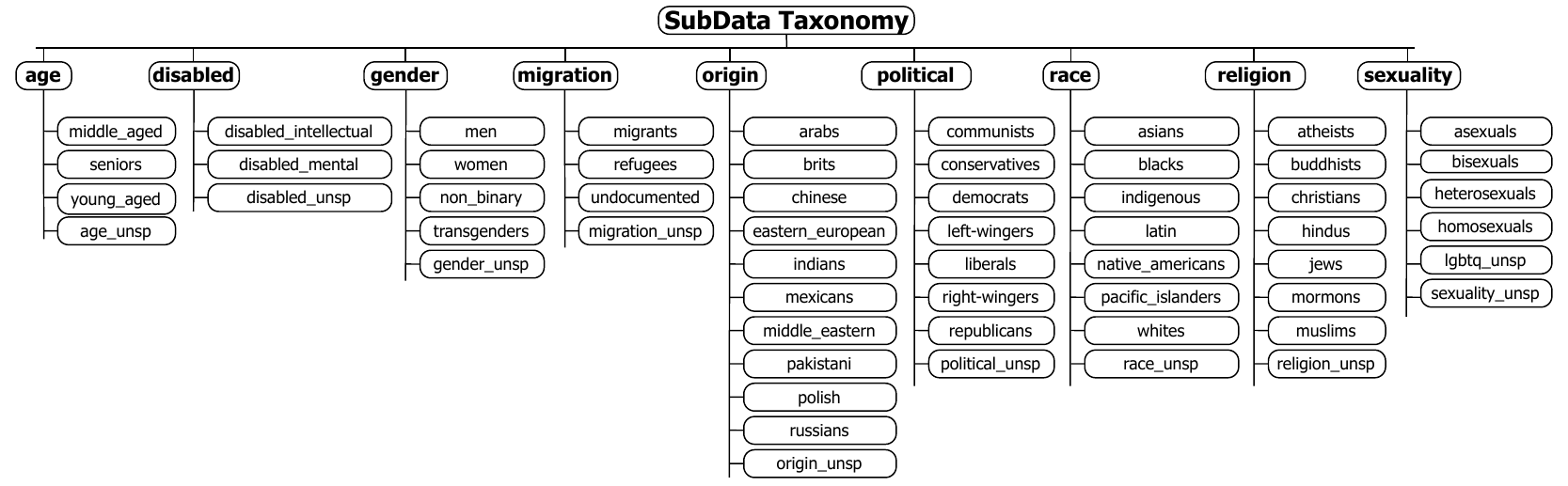}
    \caption{\textsc{SubData} taxonomy structure with target groups organized by category. \textit{Note:} targets that should end in ``\_unspecified'' have been abbreviated in the figure using ```\_unsp.''}
    \label{fig:taxonomy}
\end{figure*}

\subsection{Taxonomy Design Principles}
The development of our taxonomy was guided by several key design principles tailored to the practical needs of researchers studying perspective alignment. We sought to balance specificity and generalizability, preserving critical distinctions between target groups while establishing categories broad enough to facilitate meaningful cross-dataset analysis. For instance, the target group ``LGBTQ+'' is commonly used in the literature to encompass a wide range of minority sexual and gender identities. While we recognize that this label can be overly broad, potentially obscuring the diverse experiences of the groups it covers, we decided against introducing every identity under this umbrella as a separate target group.

Importantly, our demographic categories were not arbitrarily chosen; they emerged from a bottom-up approach, derived directly from the categories present in the original datasets we sourced. This method ensures that our taxonomy reflects and unifies the actual structure of existing hate speech research, maintaining alignment with the data's inherent organization. Additionally, whenever possible, we preserved consistency with the original researchers' taxonomic decisions to honor their methodological choices and conceptual frameworks.

\begin{table}[t]
\centering
\small
\begin{tabular}{lll}
\toprule
\textbf{Dataset} & \textbf{Original Keyword} & \textbf{Target} \\
\midrule
\citet{fanton2021human} & ``JEWS'' & jews \\
\citet{hartvigsen2022toxigen} & ``jewish'' & jews \\
\citet{jikeli2023antisemitism} & ``Kikes'' & jews \\
\citet{vidgen2021introducing} & ``jewish people'' & jews \\
\midrule
\multirow{2}{*}{\citet{vidgen2021learning}} & \multirow{2}{*}{``bla, jew''} & jews \\
    & & blacks \\
\midrule
\citet{vidgen2021learning} & ``bla, african'' & blacks \\
\citet{jigsaw2019} & ``black'' & blacks \\
\citet{jikeli2023general} & ``Blacks'' & blacks \\
\citet{rottger2021hatecheck} & ``black people'' & blacks \\
\bottomrule
\end{tabular}
\caption{Standardization of target terminology across datasets using \textsc{SubData}'s mapping system. The table provides examples of how diverse original keywords from multiple hate speech datasets are normalized into consistent target categories.}
\label{tab:mapping}
\end{table}

\subsection{Target Group Mapping}

The mapping process converts heterogeneous target labels from original datasets into our standardized taxonomy. This involves both direct equivalences (e.g., ``Jewish people'' $\rightarrow$ ``jews'') and more complex decisions requiring contextual judgment. Table~\ref{tab:mapping} provides a sample of our mapping strategy across multiple datasets, illustrating how diverse original terminology is standardized in \textsc{SubData}.

For ambiguous cases, we consulted dataset documentation to determine the original authors' intent. For instance, determining whether the target ``mexicans'' should be mapped to the ``latin'' (race category) or ``mexicans'' (origin category) required careful contextual judgment. When documentation clarified the original creators' intended meaning, we followed their categorization. When such guidance was unavailable, we applied consistent principles across similar cases.  

As part of our approach, for each category we designated target groups with the suffix ``\_unspecified'' (e.g., ``disabled\_unspecified,'' ``race\_unspecified'') to handle cases where the original dataset used generic terminology without specifying subtypes. 

Figure \ref{fig:taxonomy} illustrates the complete taxonomy structure with all target groups organized by category.

\subsection{Taxonomy Limitations and Customization}
Despite our efforts to create a comprehensive framework, we acknowledge several limitations in our taxonomy that primarily stem from the inherent challenges associated with the matching we are performing \cite{shvaiko2011ontology}. These include the LGBTQ+ target group heterogeneity that mixes gender identities and sexual orientations, blurred distinctions between racial identity and geographic origin, and simplified representations of demographic intersectionality mapped to single-attribute target groups (e.g., ``blacks,women'').
Independent from our work, \citet{fillies2025improving} point to the same challenges when developing their targeted hate speech taxonomy, relying on similar strategies to solve them.

We are confident that our taxonomy represents a useful basis for different research purposes and take the large overlap with the unified taxonomy proposed by \citet{fillies2025improving} as evidence for convergence on a generally accepted targeted hate speech taxonomy. However, recognizing that no single taxonomy can satisfy all research needs, \textsc{SubData} provides several customization functions that give researchers flexibility in adapting the framework to their specific requirements (more about it in Appendix~\ref{s:library}). While this customizability is valuable, it creates challenges for maintaining comparability across studies when researchers modify the taxonomy. To address this issue and increase transparency, we implemented a functionality to export a LaTeX version of the taxonomy (and all other modifiable resources) that researchers can include directly in their manuscripts, clearly documenting any modifications they have made.


\section{Theory-Driven Hypothesis Testing}
\label{s:theory}

The \textsc{SubData} library not only provides standardized datasets but also serves as a foundation for a theory-driven approach to evaluating LLMs perspective alignment. This approach follows the process illustrated in Figure~\ref{fig:f1}:\vspace{-0.5em}

\begin{enumerate}
    \item Theory (\textbf{T}): Researchers begin by identifying established social or political theories that predict differences in how various demographic or ideological groups differ in their perception of subjective constructs.\vspace{-0.5em}
    
    \item Hypothesis (\textbf{H}): Based on these theories, researchers formulate testable hypotheses about how LLMs aligned with different perspectives might classify content.\vspace{-0.5em}
    
    \item Experiment (\textbf{E}): Using \textsc{SubData}'s standardized datasets, researchers design controlled experiments to test these hypotheses by measuring classification differences between differently-aligned models.
\end{enumerate}

\paragraph{Advantages of the Framework.}
The theory-driven framework we propose offers substantial benefits for researchers studying LLM perspective alignment. By focusing on comparative model behavior rather than adherence to supposedly objective standards, our approach \textbf{(1) elegantly circumvents the persistent challenge of subjectivity in human annotations}. When dealing with inherently subjective constructs like hate speech, the framework does not require consensus on ``ground truth'' labels—which are often contested and vary across demographic and ideological lines—but instead directly measures differences between models aligned with distinct perspectives. This shift in evaluation methodology acknowledges the fundamental subjectivity of these tasks while still enabling rigorous analysis by grounding the tested hypotheses directly in theory.

Furthermore, our approach \textbf{(2) enables precise quantitative measurement of alignment effects on classification behavior}. Researchers can measure exactly how much perspective alignment influences model outputs when classifying content targeting specific demographics, providing concrete metrics rather than relying on qualitative assessments. This quantitative foundation makes evaluations more rigorous and facilitates meaningful comparisons across different studies, contributing to more cumulative research in this emerging field.

The framework's versatility extends beyond its primary application in political alignment evaluation. It \textbf{(3) naturally supports diverse research directions}.
This flexibility makes our approach valuable for researchers working at the intersection of natural language processing (NLP), social science, and ethical AI development, potentially informing more nuanced approaches to model development and evaluation.

\begin{table*}[t!]
\centering
\small
\rowcolors{2}{gray!10}{white} 
\begin{tabular}{l | ccT | ccT | ccT}
\toprule
\textbf{Target Category} & \multicolumn{3}{c}{\cellcolor{gray!20}\textbf{Mistral-v0.3-7B}} 
& \multicolumn{3}{c}{\cellcolor{gray!20}\textbf{Llama-3.1-8B}} 
& \multicolumn{3}{c}{\cellcolor{gray!20}\textbf{Qwen-2.5-7B}} \\
\cmidrule(lr){2-4} \cmidrule(lr){5-7} \cmidrule(lr){8-10}
 & Left & Right & \textit{OR} & Left & Right & \textit{OR} & Left & Right & \textit{OR} \\
\midrule

blacks     & \textbf{0.548} & \textbf{0.471} & 1.359*** & 0.625 & 0.614 & 1.050*** & 0.319 & \textbf{0.311} & 1.037** \\
muslims    & 0.471 & 0.391 & 1.389*** & 0.599 & 0.576 & 1.100*** & 0.228 & 0.204 & 1.152*** \\
lgbtq\_unsp & 0.313 & 0.247 & 1.389*** & 0.391 & 0.372 & 1.086*** & 0.168 & 0.157 & 1.084*** \\
jews       & 0.497 & 0.405 & 1.450*** & 0.576 & 0.553 & 1.101*** & \textbf{0.320} & 0.307 & 1.064*** \\
asians     & 0.343 & 0.260 & 1.481*** & 0.469 & 0.446 & 1.096*** & 0.197 & 0.176 & 1.146*** \\
latin      & 0.378 & 0.297 & 1.443*** & 0.473 & 0.447 & 1.110*** & 0.209 & 0.199 & 1.064*** \\
women      & 0.364 & 0.298 & 1.347*** & 0.477 & 0.467 & 1.042** & 0.161 & 0.151 & 1.073*** \\
christians & 0.202 & 0.167 & 1.263*** & 0.298 & 0.290 & 1.040** & 0.064 & 0.061 & 1.057* \\
men        & 0.299 & 0.250 & 1.279*** & 0.442 & 0.429 & 1.056*** & 0.111 & 0.104 & 1.081*** \\
whites     & 0.523 & 0.440 & 1.393*** & \textbf{0.656} & \textbf{0.649} & 1.033* & 0.259 & 0.253 & 1.029* \\
\midrule
\textbf{Overall} & 0.394 & 0.323 & 1.364*** & 0.501 & 0.484 & 1.068*** & 0.204 & 0.193 & 1.073*** \\
\bottomrule
\end{tabular}
\begin{tablenotes}
\footnotesize
\item \textbf{Note}: Significance levels are reported after correcting for multiple hypothesis testing: * $p<0.05$, ** $p<0.01$, *** $p<0.001$.
\end{tablenotes}
\caption{Hate speech detection rates by target category and persona position across the three LLMs investigated. Each cell shows the average proportion of content flagged as hateful when targeting the specified group, using a persona-conditioned model with 20 left- and 20 right-oriented personas. Odds Ratios (OR) quantify detection differences, with OR > 1 indicating higher rates for left personas. Bold values indicate the highest detection rate for each model–condition pair across all targets. Across all categories, left-oriented personas show higher detection, indicating greater sensitivity to hateful content.}
\label{tab:hate}
\end{table*}

\section{Example Use of \textsc{SubData}}
\label{sec:exp-example}
To demonstrate the proposed framework, we present here a concrete use case.  \citet{feng2023pretraining} show that pretraining LLMs on partisan corpora shifts their political leaning, and that this shift propagates into downstream tasks such as hate speech detection, where left-leaning models tend to flag more content targeting minority groups than right-leaning ones (\textbf{T}). Following their categorization (\textsc{blacks, muslims, lgbtq+, jews, latin, women, men, christian, white}), we hypothesize that a similar dynamic holds when partisan alignment is induced through persona-conditioning (\textbf{H}). Specifically, LLMs conditioned on left-leaning personas should produce higher detection rates for hate speech against minority groups, while LLMs conditioned on right-leaning personas should show the opposite tendency.
The following subsections detail the experimental setup (\textbf{E}) and results. 

\subsection{Methodology}
\paragraph{Data.} We use the \textsc{SubData} library to collect and standardize the instances of interest from existing hate speech datasets. Specifically, we rely on its unified taxonomy of ten demographic groups: \textsc{blacks, muslims, lgbtq\_unsp, jews, asians, latin, women, men, christians, whites}. For each target, we call the \texttt{create\_target\_dataset()} function, which aggregates all available instances across datasets into a consistent format with harmonized labels (refer to Appendix \ref{s:library} for more details). This procedure enables us to construct a single merged dataset that ensures comparability across groups and facilitates controlled evaluation of perspective alignment. After merging, we randomly sample 2,500 statements per target group, resulting in a balanced dataset of 25,000 instances used in our experiments.

\paragraph{Language Models.} We selected three open-source, instruction-tuned conversational LLMs for our analysis: Mistral-7B-Instruct-v0.3 \cite{jiang2023mistral7b}, Llama-3.1-8B-Instruct \cite{dubey2024llama} and Qwen2.5-7B-Instruct \cite{qwen2025qwen25technicalreport}. These models were chosen for their open-source availability and moderate parameter size (7–8B), which strikes a balance between reproducibility and diversity, allowing us to derive insights that generalize across architectures. We specifically use their conversational variants, fine-tuned for instruction following \cite{ouyang2022training}, as this aligns with our methodology: leveraging in-context prompts to condition models on different personas and evaluate their hate-speech detection behavior.

\paragraph{Experimental Setup.} To simulate partisan perspectives, we adopt the political distributions introduced by \citet{bernardelle2025political}, which map persona-conditioned LLMs across the PCT ideological space. Following the approach of \citet{civelli2025impact}, we select the 20 most left-leaning and 20 most right-leaning persona descriptions from each model distribution, yielding 40 personas in total per model. Each persona is then used as an in-context instruction to condition the LLMs for hate speech detection on the unified dataset (prompt details available in Appendix \ref{a:prompts}). For every target group, we compare the classification behavior of left- and right-oriented personas, measuring the average proportion of instances labeled as hate speech. Given 25,000 statements and 40 personas, this amounts to a total of 1,000,000 inferences per model (refer to Appendix \ref{a:resources} for more details on the resources used to run the experiments).
This setup allows us to test whether persona-induced alignment reproduces the partisan effects previously observed in pretraining-based studies.

\subsection{Results}
Table~\ref{tab:hate} reports the average proportion of hate speech detected across target groups for left- and right-oriented personas, along with odds ratios quantifying systematic differences. Several consistent findings emerge.

\paragraph{Overall persona effects.} Across all three LLMs, left-oriented personas consistently yield higher detection rates than right-oriented ones. This effect is evident not only for minority groups such as \textsc{blacks, muslims, jews}, and \textsc{lgbtq+}, but also for majority groups including \textsc{christians, men,} and \textsc{whites}. The uniformity of this effect runs counter to our hypothesis that right-leaning personas would be more protective of majority groups. Instead, persona-conditioning on the left systematically raises sensitivity to hateful content, indicating a general tightening of classification thresholds rather than selective group protection. One possible explanation is that the relatively small parameter size of these models limits their ability to adopt nuanced persona perspectives, producing broad shifts in classification behavior rather than the group-specific differences anticipated---a pattern also noted by \citet{civelli2025impact}. Investigating the precise mechanism behind this asymmetry lies beyond the scope of the present study, but future work could leverage our framework with larger-scale models to test whether the hypothesized group-specific protection emerges under more expressive architectures.

\paragraph{Variation across models.} Although the gap between left and right is consistent, its magnitude differs by model family. Mistral shows the strongest divergence (across all target groups overall OR = 1.364, $p<0.001$). By contrast, Llama exhibits the smallest persona effect (OR = 1.068, $p<0.001$), while Qwen falls in between (OR = 1.073, $p<0.001$). When considering absolute protection levels, however, a different pattern emerges: averaging overall detection rates across left- and right-conditioned personas, Llama achieves the highest baseline detection (0.493), followed by Mistral (0.359), and Qwen the lowest (0.199). While odds ratios primarily capture the models’ relative responsiveness to persona-conditioning, raw detection levels reflect their intrinsic tendency to classify statements as hateful.

\paragraph{Model-specific protection of Whites.} Across models, detection rates are generally highest for statements targeting \textsc{blacks}, making this category the most consistently protected. The main exception arises with Llama, where \textsc{whites} receive the strongest protection (0.656 under left personas), surpassing \textsc{blacks} as the top category. This unusually high value—the largest single entry in the table—may reflect the model’s U.S.-centric training distribution, where discourse around race often centers explicitly on contrasts involving \textsc{whites}. By contrast, Qwen and Mistral exhibit substantially lower absolute detection for \textsc{whites}, aligning more closely with the overall trend that prioritizes minority-group protection.

\paragraph{Remarks.} Together, the results convey two concise points. First, persona-conditioning produces an asymmetric within-model effect: left-conditioned personas yield higher hate-speech detection rates relative to right-conditioned personas, consistent with a general tightening of classification thresholds rather than selective protection of particular groups. Second, the effect’s magnitude and the models’ baseline tendencies differ: Mistral is the most responsive to persona shifts, Llama shows a higher baseline detection with notable outliers, and Qwen is comparatively stable.

\section{Conclusion}
This paper introduces a two-step framework for the systematic evaluation of perspective alignment in LLMs. First, we present \textsc{SubData}, an open-source library that standardizes heterogeneous datasets by unifying annotation schemes and demographic taxonomies, thereby enabling consistent evaluation across subjective NLP tasks. Second, we propose a theory-driven evaluation approach that leverages these standardized datasets to test hypotheses about how differently aligned models behave in downstream applications. 
We demonstrate the practical value of this framework through an experimental use case. This example illustrates how \textsc{SubData} not only provides a resource for data integration but also facilitates rigorous, theory-grounded experimentation on LLMs perspective alignment.

\paragraph{Future Extensions.}
The most immediate extension of \textsc{SubData} is the inclusion of additional datasets, both those that we may have overlooked in our initial collection as well as those that are yet to be released. In parallel, we aim to cultivate a community of researchers interested in aligning LLMs with diverse human viewpoints, which would naturally accelerate the inclusion of additional datasets.
Moreover, we plan to broaden the scope of \textsc{SubData} by introducing additional subjective constructs. Our next priority is misinformation, for which we have already compiled an initial collection of datasets that will soon be accessible through the library.
Ultimately, we intend to develop an alternative approach for evaluating LLM alignment with different human viewpoints, focusing on annotator characteristics rather than instance features. Through these initiatives, we aspire to evolve \textsc{SubData} into a comprehensive multi-construct benchmark suite for evaluating how well LLMs align with humans across various downstream tasks.

\section*{Limitations}
While the initial implementation of \textsc{SubData} focuses on hate speech detection, this narrow scope reflects the availability of suitable datasets. We chose to release the library early because alignment research is advancing rapidly but lacks standardized resources for downstream evaluation. Even in its current state, we believe \textsc{SubData} offers immediate value for studying LLM alignment with diverse perspectives.

Our unified taxonomy required pragmatic mapping choices that inevitably involve subjective judgment. Challenges include the existence of target groups in the literature that conflate targets from different categories (e.g., ``LGBTQ+'' for minority gender identities and sexual orientations), targets that are placed into different categories in different original datasets (e.g., ``mexicans'' either put into a race---latin---or an origin category) and intersectional groups (e.g., ``blacks, women''). We applied our principles carefully to balance specificity and generalizability. While the mapping process is manual and limits scalability, we argue this effort is both necessary and valuable: meaningful taxonomies for subjective constructs require domain expertise and contextual sensitivity that automated methods often miss. It is also a one-time investment with lasting benefits, and our taxonomy already aligns with independent efforts, suggesting emerging consensus. Future versions may incorporate semi-automated clustering or embedding-based methods to propose candidate mappings, with human oversight ensuring contextual validity. At the same time, \textsc{SubData} supports customization—researchers can adapt, extend, or redefine taxonomies as needed—helping to mitigate the limitations of any single framework.

Finally, the library inherits annotation errors and biases from its source datasets. \textsc{SubData} aggregates existing annotations without re-labeling or quality control, so we encourage users to verify annotation quality and consult original documentation where appropriate.

\section*{Ethical Considerations}

While \textsc{SubData} provides valuable datasets for evaluating LLMs perspective alignment, we acknowledge potential ethical concerns. The library's aggregation of hate speech datasets creates a concentrated collection of offensive content that could be misused to train hateful models or generate toxic content. 
Additionally, our framework's ability to test how differently-aligned LLMs classify content targeting specific demographics could be misused to intentionally create biased systems. We emphasize that \textsc{SubData}'s purpose is to improve evaluation transparency and understanding of perspective alignment, not to enable harmful applications. We recognize that the target groups represented in these datasets face real discrimination and harassment. Research using \textsc{SubData} should be conducted with sensitivity to the lived experiences of these communities, and findings should be communicated in ways that avoid reinforcing harmful stereotypes or creating additional psychological harm.

\bibliography{subdata}

\appendix
\section{\textsc{SubData}'s Functionalities}
\label{s:library}
While Figure \ref{fig:f1} gives an abstracted overview of the \textsc{SubData} library's basic workflow, the user-facing functionalities are documented in the following subsections. Building upon the dataset selection strategy outlined in
$\S$\ref{s:subdata} and the taxonomy and mapping strategies described in $\S$\ref{s:taxonomy}, the library offers a flexible framework that enables researchers to: \textbf{(1)} access instances targeting specific demographic groups across multiple datasets; \textbf{(2)} customize the taxonomy and mapping according to specific research needs; and \textbf{(3)} generate consistent datasets for evaluating LLM perspective alignment.

\subsection{Core Functionalities}

The library's functionality can be organized into three main categories:
\paragraph{Dataset Creation and Access}
\begin{itemize}
    \item \underline{\texttt{create\_target\_dataset()}}: Generates a dataset containing instances targeting the specified valid target group (e.g., ``jews'', ``blacks'') from all available datasets.
    Returns a dataframe with instance ID, text, target name, and source dataset.
    
    \item \underline{\texttt{create\_category\_dataset()}}: Assembles instances targeting all groups within a specified category (e.g., ``religion,'' ``race''). 
    Downloads and processes all datasets containing any target groups within the specified category.
    
    \item \underline{\texttt{get\_target\_info()}}: Displays available instances for specific target groups, showing distribution across datasets and availability status. 
    Displays the total number of instances available, lists source datasets with counts, and provides access requirement information for restricted datasets.
    
    \item \underline{\texttt{get\_category\_info()}}: Provides an overview of available instances for all groups within a category. Displays total instance counts across all target groups in the category, breaks down counts per target, and shows dataset availability information.
\end{itemize}

\paragraph{Taxonomy Customization}
\begin{itemize}
    \item \underline{\texttt{show\_taxonomy()}}: Displays and exports the specified taxonomy. Either returns the full taxonomy or only the  specified categories (either ``all'' or a list of category names). When LaTeX export is enabled, the function generates formatted tables in a txt file, making it convenient to include taxonomy details in academic papers.

    \item \underline{\texttt{update\_taxonomy()}}: Reorganizes target groups across categories or creates new categories. This function accepts a dictionary of taxonomy changes (specifying which targets to move from which categories to which new categories) and stores the modified taxonomy under the provided name. If a target is moved to a non-existent category, a new category will be created automatically, allowing for flexible taxonomy extension.
    
    \item \underline{\texttt{add\_target()}}: Creates entirely new target groups when needed. This function requires three parameters: the name of the new target, the existing category to place it in, and a list of original dataset keywords that should map to this new target. Stores the modified taxonomy and mapping under the provided names.
    
    \item \underline{\texttt{update\_overview()}}: Updates the internal dataset overview that informs the dataset creation and information functions. This function should be called after any taxonomy or mapping modifications to ensure that future calls of the dataset-generating functions access the correct resources.
\end{itemize}

\paragraph{Mapping Modification}
\begin{itemize}
    \item \underline{\texttt{show\_mapping()}}: Displays and exports the specified mapping between original dataset keywords and standardized target groups. Either returns the individual mappings for all datasets or only for those specified (either ``all'' or a list of dataset names). The LaTeX output consists of separate tables for each dataset, clearly documenting the keyword-to-target transformations used in the research pipeline.

    \item \underline{\texttt{update\_mapping\_specific()}}: Modifies mappings for individual datasets, allowing dataset-specific customization of how original dataset labels map to standardized target groups. This function accepts a nested dictionary specifying which keywords in which datasets should be mapped to which target groups. Stores the modified mapping under the provided name.
    
    \item \underline{\texttt{update\_mapping\_all()}}: Applies mapping changes consistently across all datasets, ensuring uniform treatment of keywords. This function takes a dictionary mapping original keywords to new target groups, affecting all datasets where those keywords appear. It stores the modified mapping under the provided name.
\end{itemize}

\paragraph{Dataset Overview}

\begin{itemize}
    \item \underline{\texttt{update\_overview()}}: Updates the internal dataset overview that informs the dataset creation and information functions. This function should be called after any taxonomy or mapping modifications and accepts parameters for naming the modified configurations (overview\_name, mapping\_name, taxonomy\_name), as well as an optional authentication token (hf\_token) for accessing restricted datasets.
    
    \item \underline{\texttt{show\_overview()}}: Displays and exports the specified overview based on the specified taxonomy. This function accepts an overview name and taxonomy name as parameters, with boolean options to control JSON export (export\_json) and LaTeX table export (export\_latex). When LaTeX export is enabled, the function generates formatted tables in a txt file. The function returns the overview as a dictionary.
\end{itemize}

\subsection{Implementation and Availability}
All code is available open-source on GitHub\footnote{\url{https://github.com/Subdata-Library/Subdata/}} and the library can be installed directly from PyPi\footnote{\url{https://pypi.org/project/subdata/}}. The library handles dataset availability transparently—if a dataset is not openly available, the functions inform users how to access it, either by providing authentication credentials or manually downloading and storing datasets in a specified location.

\section{Prompt Template}
\label{a:prompts}
Across all LLMs investigated, we employed a standardized prompt format to ensure comparability of results and to minimize confounding effects introduced by variations in wording or structure. We wrapped all inputs in the respective chat template of each model family. This preserved the conversational format they were optimized for during fine-tuning, reducing the risk of degraded performance or altered behavior due to mismatched prompting styles. For each combination of persona and statement (40 personas × 25{,}000 statements = 1{,}000{,}000 total pairs), we generated the corresponding prompt by substituting \texttt{[TEXT]} and \texttt{[PERSONA]} placeholders in the template with the actual text.

To improve reliability and maintain consistent output formats, we adopted a structured output generation strategy throughout our experiments.\footnote{We used the implementation from vLLM (\url{https://docs.vllm.ai/en/v0.8.5.post1/features/structured_outputs.html}), though other toolkits offer equivalent functionality.} This approach constrains model generations to follow a predefined schema, thereby helping to prevent ill-formed or off-task completions. At inference time, schema adherence is enforced through dynamic vocabulary masking: at each decoding step, only tokens that keep the partial output consistent with the schema remain available for selection. This ensures that final outputs are both syntactically valid and semantically aligned with the intended task requirements. This design mitigated refusal behaviors and promoted consistent formatting across different models, addressing reproducibility issues observed in earlier work \cite{rottger-etal-2024-political, azzopardi2024prism}.

\begin{tcolorbox}[
    colback=gray!10!white, 
    colframe=gray!75!black, 
    title=\textbf{User Prompt}, 
    fonttitle=\bfseries, 
    boxrule=1pt, 
    width=\linewidth
]
\small
You are an expert content analyst specializing in identifying hate speech according to specific guidelines. Your task is to analyze the provided text strictly from the perspective of \textbf{[PERSONA]}.

\medskip

Based on your analysis, you must generate a single, valid JSON object that adheres to the following schema and logic. \textbf{Do not output any text other than the JSON object.}

\medskip

\textbf{Analysis Logic:}
\begin{enumerate}
    \item \textbf{Perspective:} Embody the given persona. Your entire analysis must reflect their likely values, biases, worldview, and sensitivity level to hateful content.
    \item \textbf{Classification:} Determine if the persona would consider the text to be hate speech. This determines the value of \texttt{is\_hate\_speech}.
\end{enumerate}

\medskip

\textbf{JSON Schema:}
\begin{itemize}
    \item \texttt{is\_hate\_speech} (\textit{string}): Must be one of \texttt{"true"} or \texttt{"false"}.
\end{itemize}

\medskip

\textbf{Example 1 (Hateful):}
\begin{verbatim}
{
    "is_hate_speech": "true"
}
\end{verbatim}

\textbf{Example 2 (Not Hateful):}
\begin{verbatim}
{
    "is_hate_speech": "false"
}
\end{verbatim}

\medskip

Persona: \textbf{[PERSONA]} \\
Text: \textbf{[TEXT]}

\medskip

Now, analyze the text from the persona's perspective and generate the JSON object.
\end{tcolorbox}

\section{Computational resources}
\label{a:resources}
All experimental conditions were executed on a single H100 GPU. Each run required approximately one and a half hours to complete, resulting in a total runtime of roughly 5 hours across the full set of experiments.

\end{document}